\pdfoutput=1
\documentclass[11pt]{article}

\usepackage[]{acl}

\usepackage{times}
\usepackage{latexsym}

\usepackage[T1]{fontenc}
\usepackage[utf8]{inputenc}

\usepackage{microtype}

\usepackage{inconsolata}

\usepackage{graphicx}

\title{``Why'' Has the Least Side Effect on Model Editing}

\author{Tsung-Hsuan Pan,\textsuperscript{\rm 1}
    Chung-Chi Chen,\textsuperscript{\rm 2}
    Hen-Hsen Huang,\textsuperscript{\rm 3}
    Hsin-Hsi Chen \textsuperscript{\rm 1} \\
   \textsuperscript{\rm 1} Department of Computer Science and Information Engineering, \\ 
   National Taiwan University, Taiwan \\
    \textsuperscript{\rm 2} AIST, Japan \\
    \textsuperscript{\rm 3}
    Institute of Information Science, Academia Sinica, Taiwan \\
    b08902138@ntu.edu.tw,
    c.c.chen@acm.org,\\
    hhhuang@iis.sinica.edu.tw,
    hhchen@ntu.edu.tw}

\begin{document}
\maketitle
 \begin{abstract}
Training large language models (LLMs) from scratch is an expensive endeavor, particularly as world knowledge continually evolves. To maintain relevance and accuracy of LLMs, model editing has emerged as a pivotal research area. While these methods hold promise, they can also produce unintended side effects. Their underlying factors and causes remain largely unexplored. This paper delves into a critical factor—question type—by categorizing model editing questions. Our findings reveal that the extent of performance degradation varies significantly across different question types, providing new insights for experimental design in knowledge editing.
Furthermore, we investigate whether insights from smaller models can be extrapolated to larger models. Our results indicate discrepancies in findings between models of different sizes, suggesting that insights from smaller models may not necessarily apply to larger models. Additionally, we examine the impact of batch size on side effects, discovering that increasing the batch size can mitigate performance drops.
\end{abstract}

\section{Introduction} 
\label{sec:intro}

Training large language models (LLMs) from scratch is prohibitively expensive when world knowledge changes. However, the world evolves daily. To keep LLMs updated with current world knowledge, model editing~\cite{mitchell2022memory,chen2024robust,hartvigsen2024aging,yu2024melo} has emerged as a crucial research area in the LLM era. 
Although model editing methods show potential in updating knowledge, partially modifying the parameters of language models via model editing is akin to performing surgery on the human brain, potentially leading to side effects~\cite{hoelscher-obermaier-etal-2023-detecting,gu2024model,yang2024butterfly}. 
While there are some intuitive discussions on the side effects of model editing, identifying the factors and causes of these side effects is scarcely addressed. 
We noticed that the question-answering setting is the most common when editing knowledge. For example, when we want to update the information about the U.S. president, we typically design a question for models such as ``Who is the president of the U.S.?'' Following this line of thought, we are curious whether different question types will lead to different side effects after editing.

A common finding regarding the side effects of model editing is that the model's performance across different aspects tends to deteriorate after a few edits~\cite{gu2024model,yang2024butterfly}. 
Given that the severity of surgical side effects varies with the type of surgery, we are curious whether editing the knowledge for different question types will result in varying degrees of performance degradation. 
To this end, we categorize the questions used for model editing into eight types: who, what, when, where, which, why, how, and others. 
Our results indicate that the extent of performance degradation significantly differs after editing knowledge for different types of questions. It suggests future directions for experimental design in knowledge editing. 

Moreover, if the illness issues are related or addressing them together can reduce the overall surgical risk, doctors might choose a single surgery to solve multiple problems. Based on this concept, we discuss the side effects under different batch size settings. Our results suggest that enlarging the batch size, i.e., editing several pieces of knowledge at the same time, can mitigate the side effects of the performance drop.

Finally, performing the same surgery on adults and children may result in different side effects, and the underlying causes may vary. 
Following this line of thought, we experiment with GPT-2 (1.5 billion parameters)~\cite{radford2019language} and LLaMA-7B (7 billion parameters)~\cite{touvron2023llama} to explore whether findings from smaller models, which is cheaper and more efficient, can be extrapolated to larger models. 
Unfortunately, our results indicate that the findings differ between models of different sizes, suggesting that insights from smaller models may not necessarily apply to larger models.

In sum, this paper makes the following contributions:
(1) We provide an in-depth analysis of how different question types affect the performance of LLMs after model editing.
(2) We investigate the impact of batch size on the side effects of model editing and reveal that larger batch sizes can mitigate performance degradation.
(3) We explore the applicability of findings from smaller models to larger models and highlight the limitations of such applications.

\section{Related Work}
Model editing is a rapidly evolving field with several key approaches aimed at modifying model behavior without extensive retraining~\cite{Yao2023EditingLL}. Fine-tuning with constraints~\cite{Zhu2020ModifyingMI} is a method developed to mitigate the issue of catastrophic forgetting, where new knowledge overwrites previously learned information. This approach involves updating as few parameters as possible or only modifying specific parts of the model's structure. Memory-augmented techniques~\cite{Mitchell2022MemoryBasedME} involve storing new or corrected knowledge separately from the original model, effectively creating a patch model. These patches can be implemented in various ways, such as through pretrained models or datastores, and are combined with the original model using simple methods like classifiers. However, this approach requires retraining both the classifier and the patch model, which is not ideal for continuous updates. 
Hyper networks~\cite{DeCao2021EditingFK} represent a dynamic method where the model continuously updates its parameters based on incoming knowledge without needing retraining or fine-tuning. This is achieved by training a network to predict the weights of another network, effectively learning the process of fine-tuning through gradient descent. Despite its promise, the efficacy of hyper networks may diminish as the volume of updates increases, posing challenges for long-term usability. Additionally, current implementations can handle only up to 75 knowledge edits at a time.

The locate-and-edit approach~\cite{Meng2022LocatingAE,Meng2022MassEditingMI} leverages interpretability insights, treating the MLP layers in transformers as key-value memories~\cite{Geva2020TransformerFL}. By identifying the specific neurons responsible for storing factual associations (keys and values), this method modifies the values corresponding to the desired knowledge. The process involves evaluating the influence of individual neurons on the output and adjusting the most impactful ones. It offers enhanced interpretability and allows for precise targeting of specific pieces of knowledge within the model. It is favored for scenarios where understanding and precisely controlling model behavior is crucial.
Therefore, in this paper, we focus on the iconic method of locate-and-edit, MEMIT~\cite{Meng2022MassEditingMI}, for in-depth analysis and discussions.

\begin{figure*}
    \centering
    \resizebox{\textwidth}{!}{
    \includegraphics{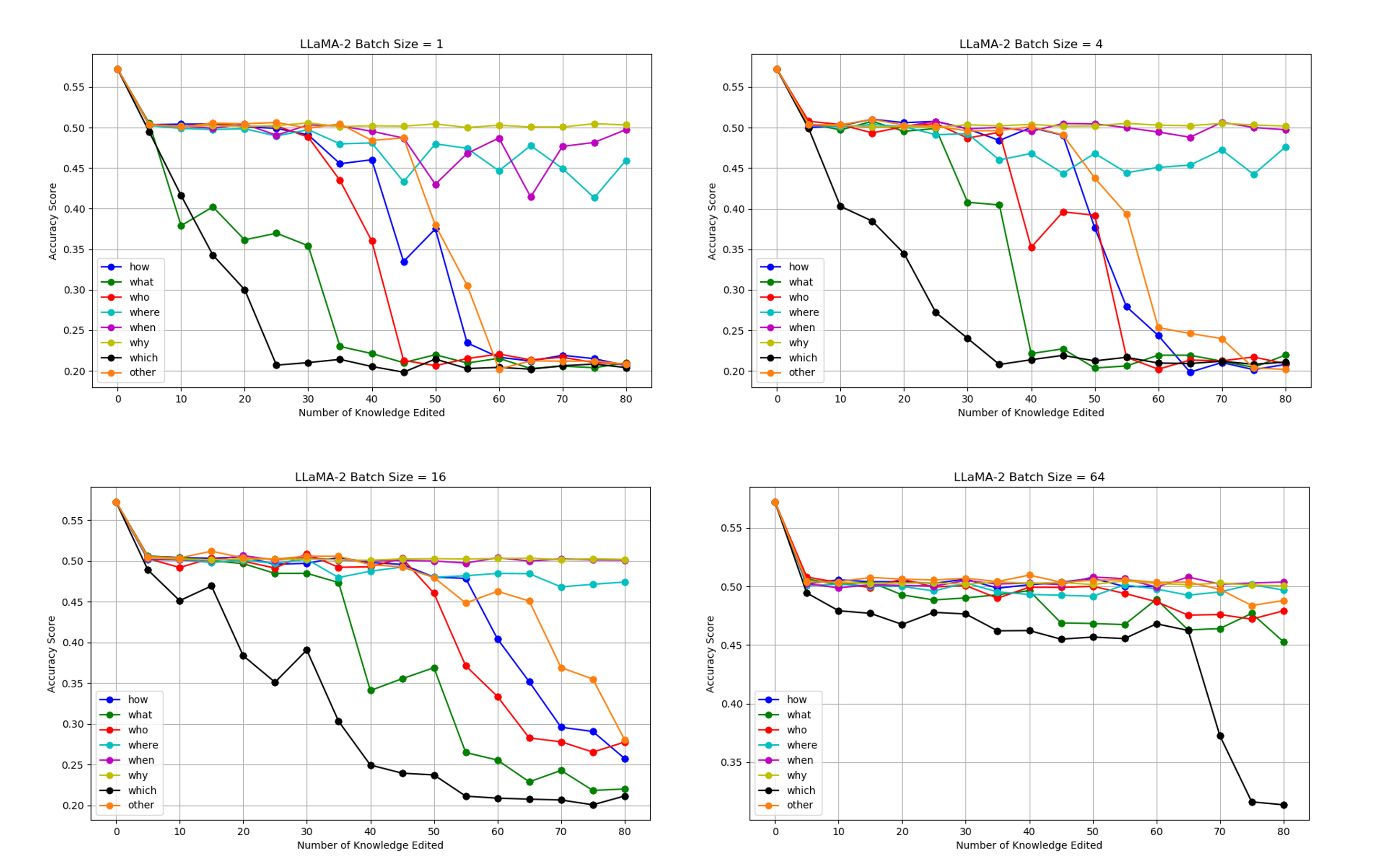}
    }
    \caption{Results of LLaMA-2. Please note that the scale of the y-axis in different charts differs for the detailed discussions.
}
    \label{fig:llama2}
\end{figure*}

\section{Experimental Setup}
\subsection{Knowledge Editing Dataset}
We use RealTimeQA~\cite{Kasai2022RealTimeQW} as the base dataset for knowledge editing. RealTimeQA is a collection derived from popular news sources, containing articles from various news websites. Weekly, RealTimeQA gathers news articles along with approximately 30 multiple-choice questions authored by humans from platforms such as CNN, THE WEEK, and USA Today, covering diverse topics including politics, business, sports, and entertainment. 
Unlike other datasets such as ZsRE~\cite{Levy2017ZeroShotRE} or CounterFact~\cite{Meng2022LocatingAE}, which draw from known Wiki knowledge or focus on false facts respectively, we opt for RealTimeQA due to its alignment with real-world scenarios, offering a more fitting context for our knowledge updating needs. 
In our experiment, we randomly selected 80 questions of each question type from a total of 1,781 instances.

\subsection{General Ability Evaluation}
To assess the model's general ability, including knowledge acquisition, comprehension, and reasoning abilities, we utilize ARC-easy, ARC-challenge~\cite{Clark2018ThinkYH}, and OpenBookQA~\cite{Mihaylov2018CanAS} as our primary evaluation datasets. The ARC Benchmark, featuring over 7,787 science questions spanning from 3rd to 9th-grade standardized test levels, presents formidable challenges for both retrieval-based and word co-occurrence algorithms, particularly in its Challenge Set. This division into Easy and Challenge Sets allows for a nuanced examination of performance across varying difficulty levels. Additionally, OpenBookQA introduces a novel evaluation paradigm inspired by open-book exams, demanding a profound understanding of elementary-level science facts and their practical application in diverse scenarios. Through these datasets, we aim to comprehensively evaluate our model's capabilities across varying levels of complexity and real-world applicability, from basic knowledge retrieval to sophisticated reasoning tasks.

\subsection{Evaluation Paradigm}
We chose to experiment with GPT-2-XL (1.5B)~\cite{Radford2019LanguageMA} and LLaMA-2 (7B)~\cite{Touvron2023Llama2O} as our testing models to explore the impact of model size on performance outcomes. GPT-2-XL represents a mid-sized model, while LLaMA-2 is substantially larger, allowing us to observe potential trade-offs between computational efficiency and performance gains.
To discuss the side effects of model editing, we use MEMIT~\cite{Meng2022MassEditingMI} to edit models based on the knowledge changes in RealTimeQA with different types of questions and different settings on the batch size. Then, we test the models' general ability with ARC-easy, ARC-challenge, and OpenBookQA, and report the average accuracy as the evaluation for general ability.

\begin{figure*}
    \centering
    \resizebox{\textwidth}{!}{
    \includegraphics{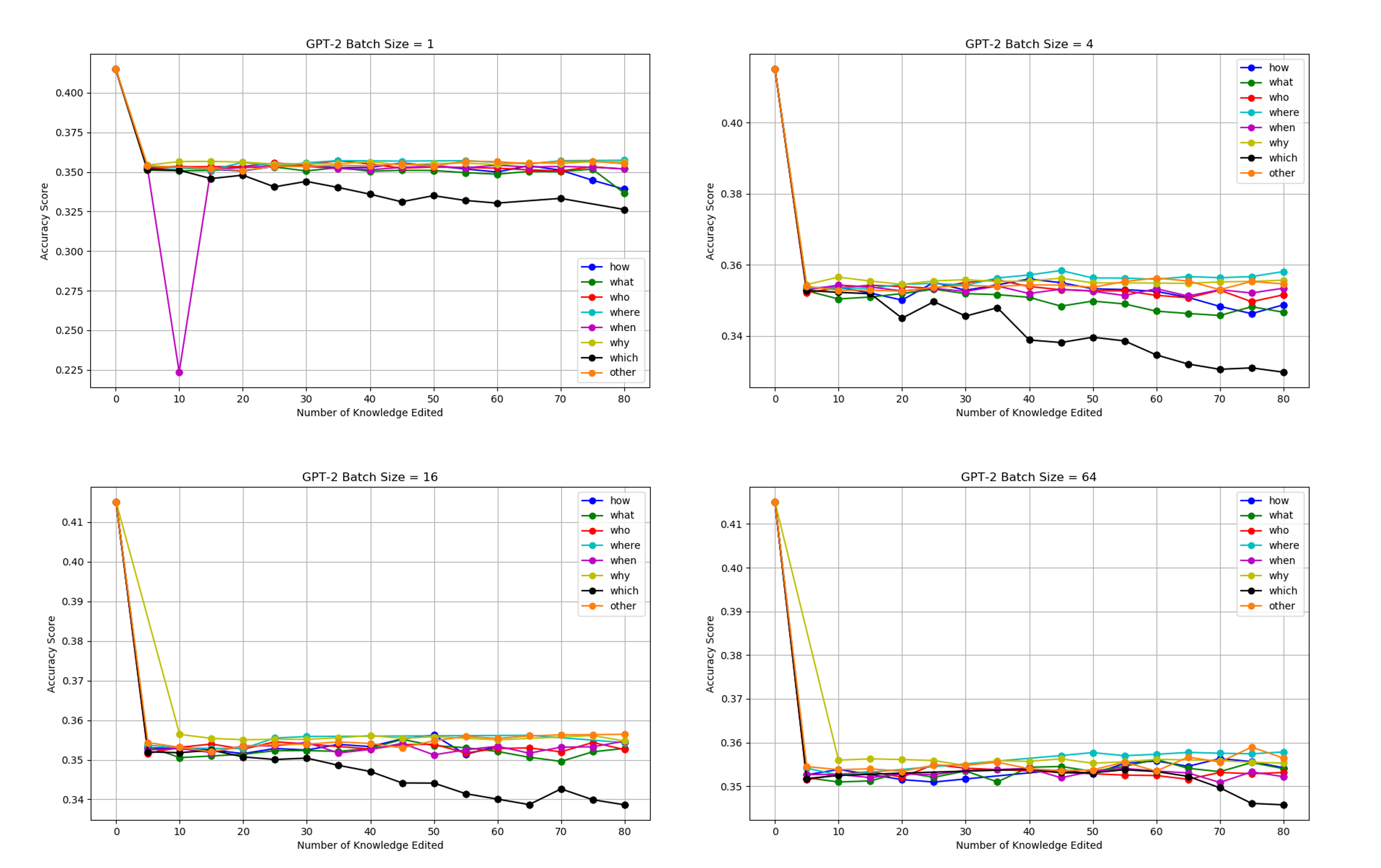}
    }
    \caption{Results of GPT-2.}
    \label{fig:gpt2}
\end{figure*}

\section{Results and Analysis}
\subsection{Impact of Question Type}

Figure~\ref{fig:llama2} illustrates the general ability of LLaMA-2 7B as the number of knowledge edits increases under different batch size settings. 
We first examine the results for a batch size equal to 1 (upper left subfigure in Figure~\ref{fig:llama2}). 
The results reveal a significantly different trend in the model's performance after editing knowledge based on different question types. 
For all question types, the general ability drops to around 50\% after five knowledge edits. 
This finding is consistent with previous studies~\cite{gu2024model,yang2024butterfly}, indicating that a few edits can lead to model collapse. 
However, a deeper analysis of this side effect shows that after editing 10 knowledge items, the general ability drops significantly more for ``which'' or ``what'' questions, while the general ability for other question types remains stable. 

Furthermore, as the number of knowledge edits increases, the general ability of the model edited under different question types drops sequentially rather than simultaneously. 
These results suggest that different question types have varying impacts on the model's general ability. 
Notably, ``Why'' questions have the least adverse effect on model editing. 
The general ability of the model edited with ``Why'' questions does not drop a second time, unlike other question types. 
We hypothesize that this is because LLMs are trained for continuous writing, and answers to ``Why'' questions are full sentences, whereas answers to other questions mainly involve editing named entities. 
For instance, ``where'' questions edit knowledge related to locations, and many ``how'' questions are about quantities, such as ``how many'' and ``how much''.

\subsection{Mitigating Side Effects}

We compare the effects under different batch size settings in Figure~\ref{fig:llama2}. 
Under varying batch sizes, the observations of the performance drop across different question types are similar, including the second drop and the order of dropping among different question types. 
However, we observed that the timing of the second drop is delayed as the batch size increases. 
These results suggest that editing the same type of questions simultaneously may help mitigate side effects.

\subsection{Observations on Model Size}

As mentioned in Section~\ref{sec:intro}, experimenting with LLMs is more expensive and time-consuming than with smaller language models. 
We conducted the same experiments with GPT-2, and the results are shown in Figure~\ref{fig:gpt2}. 
Although there are some minor fluctuations, the general ability drops to the lowest level directly without a second drop, regardless of the question types and batch size. 
These results indicate that the side effects and observations with smaller language models may differ from those with large language models. 
It also suggests that the behaviors of these two types of models should be considered and analyzed independently, despite the side effects occurring in both after a few edits.

\section{Conclusion}
This paper explored the factors influencing the side effects of model editing. Our findings highlight the significant role of question type in determining the extent of performance degradation, revealing that ``why'' questions have the least adverse effect on model stability. 
We also examined the impact of batch size on side effects, discovering that increasing the batch size can delay performance drops. 
Lastly, our comparison between GPT-2 and LLaMA-2 7B revealed that insights from smaller models do not always extrapolate to larger ones. This discrepancy highlights the need for independent analysis of different model sizes, emphasizing that findings from smaller models should be cautiously applied to larger LLMs.

\section*{Limitation}
First, our study focuses on eight specific question types. This categorization, while comprehensive, may not cover all possible variations of model queries encountered in real-world applications. Future work could explore additional question types or more nuanced classifications to provide a broader understanding of the impact of question types on model editing.
Second, we conducted our experiments on two specific models: GPT-2 and LLaMA-7B. The discrepancies observed between these models highlight the need for caution when generalizing findings to other models.
Third, our assessment focused on the general ability of models post-editing. However, other important metrics, such as interpretability and robustness, were not considered. Including these metrics in future studies could offer a more holistic view of the consequences of model editing.
Finally, while we identified different impacts of question types and batch sizes on model performance, the underlying mechanisms driving these side effects remain unclear. Further research is needed to understand the causal relationships and develop methods to predict and mitigate unintended consequences effectively.

\bibliography{custom}

\end{document}